\documentclass[journal]{IEEEtran}
\ifCLASSINFOpdf
\else
\fi
\usepackage{amssymb}
\usepackage[noadjust]{cite}
\usepackage{graphicx}
\usepackage{csquotes}
\usepackage{amsmath}
\usepackage{hyperref}
\usepackage{multirow}
\usepackage{cellspace}
\usepackage{subfig}

\begin{document}
%
\title{Metric-Learning based Deep Hashing Network for Content Based Retrieval of Remote Sensing Images}
%
%
%

\author{Subhankar~Roy,~\IEEEmembership{Student Member,~IEEE,}
        Enver~Sangineto,
        Beg\"{u}m~Demir,~\IEEEmembership{Senior Member,~IEEE,} and~Nicu~Sebe,~\IEEEmembership{Senior Member,~IEEE}
}
\maketitle

\begin{abstract}
Hashing methods have  recently been shown to be very effective in retrieval of Remote Sensing (RS) images due to their computational efficiency and fast search speed. Common hashing methods in RS are based on hand-crafted features on top of which they learn a hash function which provides the final binary codes. However, these features are not optimized for the final task (i.e., retrieval using binary codes). On the other hand, modern Deep Neural Networks (DNNs) have shown an impressive success in learning optimized features for a specific task in an end-to-end fashion. Unfortunately, typical RS datasets are composed of only a small number of {\em labeled} samples, which makes the training (or fine-tuning) of big DNNs problematic and prone to overfitting.
To address this problem, in this paper we introduce a metric-learning based hashing network, which: 1) Implicitly uses a big, pre-trained DNN as an intermediate representation step without the need of re-training or fine-tuning; 2) Learns a semantic-based metric space where the features are optimized for the target retrieval task;  3) Computes compact binary hash codes for fast  search. 
Experiments carried out on two RS benchmarks highlight  that the proposed network significantly improves the retrieval performance under the same retrieval time when compared to the state-of-the-art hashing methods in RS.
\end{abstract}


%
\IEEEpeerreviewmaketitle

\vspace{-5mm}
\section{Introduction}
\label{Introduction}
%
%
%
%
\IEEEPARstart{T}{he} advancement of satellite technology has resulted in an explosive growth of Remote Sensing (RS) image archives.  Content-Based Image Retrieval (CBIR) methods in RS aim  to retrieve those archive images which are  the most similar with respect to a given query image.
Traditional Nearest Neighbour (NN) search algorithms 
(which exhaustively compare the query image with all the images in the archive) 
are time consuming and thus prohibitive for  large-scale RS image retrieval problems \cite{NN}. To alleviate this issue, hashing-based  search techniques have been recently proposed in RS. A hashing-based method involves learning a hash function, $\mathbf{b} = h(\mathbf{d})$, which maps a high-dimensional image descriptor $\mathbf{d} \in {\rm I\!R}^{D}$ into a  low-dimensional binary code $\mathbf{b} \in \{0,1\}^{\textit{K}}$, where $D >> \textit{K}$. Besides improving the computational efficiency, hashing methods also reduce the storage costs significantly. For instance,  Locality-Sensitive Hashing (LSH) \cite{slaney2008locality}  learns $n$ different hash functions $[h_1, h_2,..., h_n]$ by choosing $n$ random projecting vectors obtained from a multivariate Gaussian distribution. 
In the context of CBIR in RS, two hashing-based methods have been introduced in \cite{demir2016hashing, reato2019unsupervised, li2018large}. Demir et al. \cite{demir2016hashing}  learn two hash functions in the kernel space using hand-crafted features (SIFT) and a bag-of-visual-words  representation.  \cite{reato2019unsupervised} extends these hashing methods by describing each image with multi-hash codes and
using the spectral histograms of the image regions. However, the effectiveness of the binary codes relies on both the approach used for the hash-function  and  the adopted  image descriptors. Specifically, common hand-crafted features are not optimized for the retrieval task and have a limited capability to accurately represent the high-level semantic content of RS images. 

The limitations of hand-crafted features is nowadays addressed by using  deep Convolutional Neural Networks (CNNs), which learn a feature space directly optimized for the final task (e.g., classification). For instance, Li et al., \cite{li2018large} introduce a deep Hashing Neural Network (DHNN) to address CBIR in RS.  DHNN jointly learns semantically accurate deep  features and binary hash codes, by means of a cross-entropy loss  function. However,  entropy-based losses are particularly effective in classification problems, but less effective in defining a {\em metric space} which clusters together  similar  images, a concept which is crucial in a CBIR framework. Specifically, the absence of a margin threshold between positive and negative samples leads to a poor generalization. As a result, to achieve a high retrieval performance,  DHNN requires long hash codes and a large amount of {\em annotated} training images, which is difficult to collect in common RS archives.

To address these issues, we propose to use an intermediate representation, obtained with a big CNN (Inception Net \cite{szegedy2016rethinking}) pre-trained on ImageNet and {\em not} fine tuned. In our approach Inception Net is not re-trained, thus we avoid overfitting risks when using  small labeled datasets, as those commonly available in RS. Inception Net is pre-trained on ImageNet, a dataset of more that one million (labeled), non-RS images. The visual knowledge acquired by Inception Net on this dataset is exploited to extract an intermediate representation of our RS images. Using this representation as input, we {\em train} a second network  whose  final features are jointly optimized (using a combination of different losses) for both the retrieval task and the  hash-code production. 
In summary, the contributions of
our   \textbf{M}etr\textbf{i}c-\textbf{L}earning based Deep H\textbf{a}shing \textbf{N}etwork (MiLaN)
 are: (1) We use as input an intermediate feature representation  to alleviate overfitting problems on small RS annotated datasets.
 (2) We train  a DNN  on top of these intermediate features, whose final feature representation is   jointly optimized  for both a CBIR task and a hash-code production task.  
This letter extends our work presented in \cite{roy2018deep} introducing a detailed ablation study of 
MiLaN, a latent-space visualization and additional experiments, including testing on another RS dataset.

\vspace{-2mm}
\section{Metric-Learning based Deep Hashing Network} 
 We assume that a small set of training images is available:  $\mathcal{I} = \{x_1, x_2, ..., x_P\}$, where each $x_i$ is associated with a corresponding class label $y_i \in \mathcal{Y} = \{y_1, y_2, ..., y_P\}$.
Our goal is  to learn a hash function  encoding an image into a binary code: $h : \mathcal{I} \rightarrow \{0,1\}^K$, where $\textit{K}$ is the number of bits in the hash code. Importantly,  $h$  should capture the semantics of the images in a similarity space.
This  is achieved using a specific loss function which clusters together  semantically similar samples. Specifically, we use  a \textit{triplet loss} \cite{schroff2015facenet}, which learns a metric space such that the  Euclidean distance between two points in this  space faithfully corresponds to the visual similarity  between the corresponding pair of images in the pixel space. Moreover, we use two additional losses: 1) A representation penalty loss, which pushes the activations of the last layer of  the network to be binary; and 2) A bit-balancing loss, which encourages the network to produce hash codes having (on expectation) an equivalent number of 0s and 1s. The latter is  important to guarantee that all the final bit codes are effectively useful for the binary-based retrieval task.

\begin{figure}[h!]
	\centering
	\includegraphics[width=0.9\columnwidth]{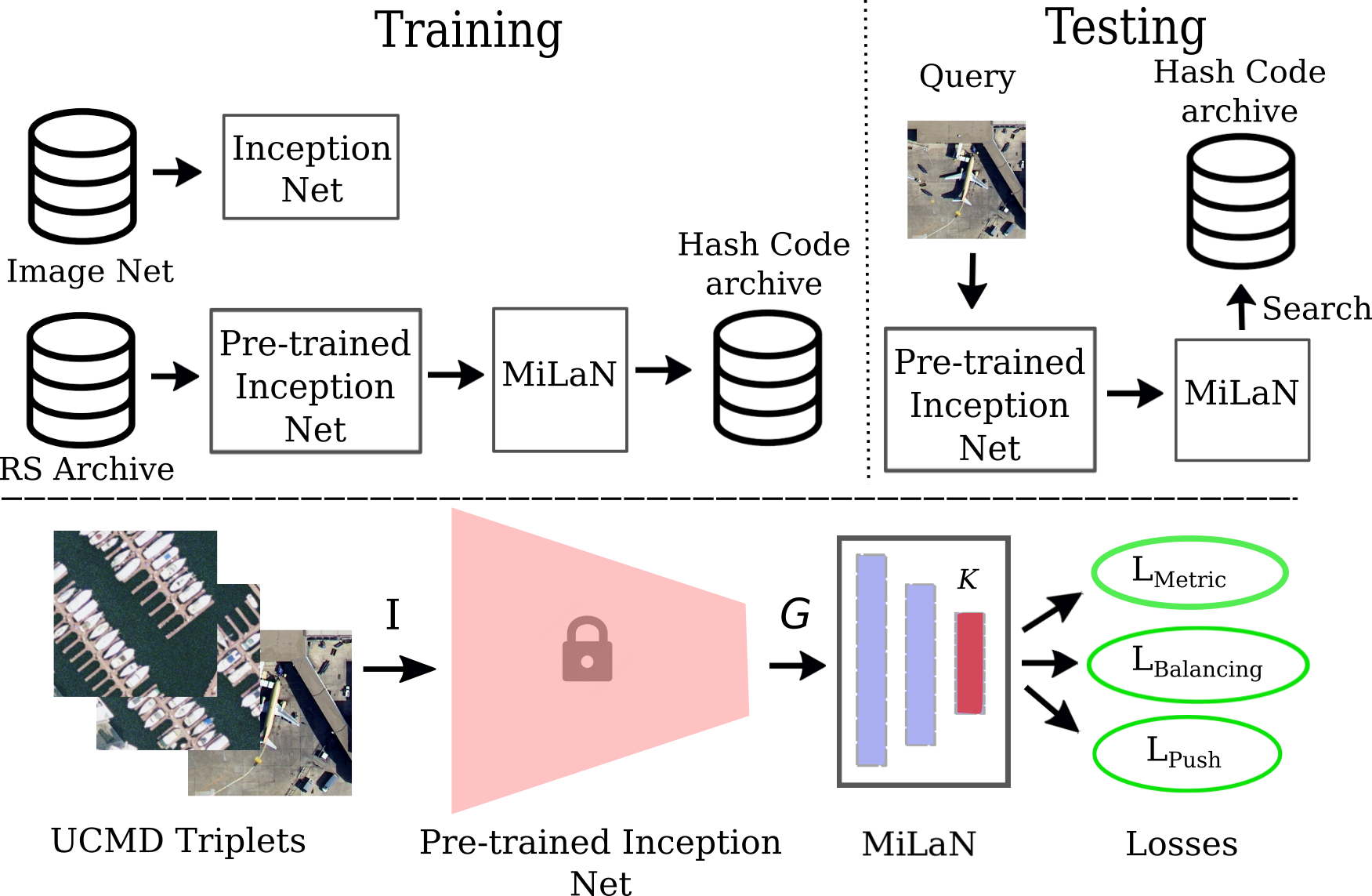}
	\caption{Top: Overall pipeline. Bottom: Inception Net is used to extract an intermediate  image representation, which is then fed to our MiLaN to obtain binary codes of length \textit{K}.}
	\label{fig:net}
\end{figure}

As mentioned in Sec. \ref{Introduction}, the success of DNNs in vision tasks relies on the use of big CNNs which extract visual information from raw-pixel data. However, these big networks are prone to overfit when trained using small labeled dataset, which is the common case in RS frameworks. In order to solve this problem, we propose to use a big CNN  (Inception Net \cite{szegedy2016rethinking}) pre-trained on ImageNet, without re-training. Specifically, Inception Net is used to extract an intermediate representation of our RS images. Then, a smaller network ($f$) is trained using this representation as input in order to learn our final hash function.
In more detail, each image in $\mathcal{I}$ is fed to the Inception Net \cite{szegedy2016rethinking} and we use the last layer ($Pool_3$)  right before the softmax layer as our intermediate feature representation. This layer is composed  of 2048 neurons, whose activations represent the result of an average pooling on the convolutional feature maps of the layer before (Fig.\ref{fig:net}). 
 
\begin{figure}[h!]
	\centering
	\includegraphics[width=0.9\columnwidth]{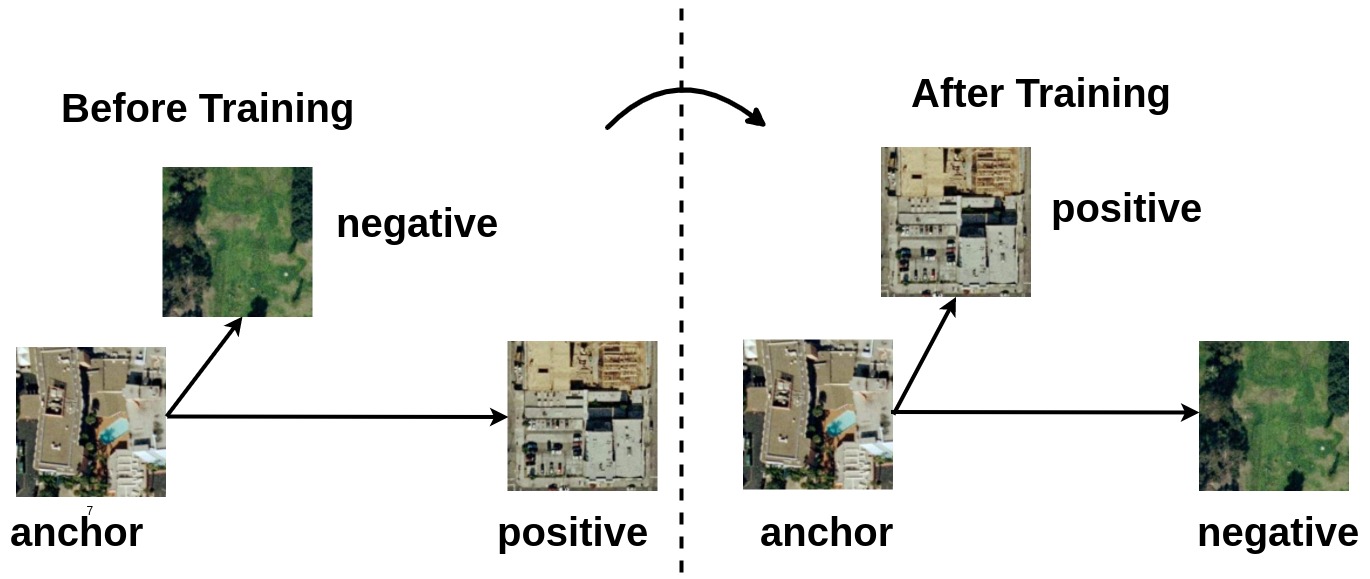}
	\caption{The intuition behind the triplet loss: after training, a positive sample is \enquote{moved} closer to the anchor sample than the negative samples of the other classes.}
	\label{fig:triplet_loss}
\end{figure}

Let $\mathcal{G}=\{\mathbf{g}_1, \mathbf{g}_2,..., \mathbf{g}_P\}$, $\mathbf{g}_i \in {\rm I\!R}^{2048}$ be  the set of the intermediate features  extracted from  $\mathcal{I}$. The elements of $\mathcal{G}$ are fed to the  network $f$, whose  weights are  {\em randomly initialized}. The goal of $f$  is to learn a mapping from the intermediate features to a metric space which is semantically significant for the specific RS retrieval task: $f : {\rm I\!R}^{2048} \rightarrow {\rm I\!R}^K$. The real-valued activations of the last layer of $f$  are finally quantized using a simple thresholding to obtain the  binary hash codes. 

The main  loss used to train $f$  is the triplet loss, which is based on the intuition that images with the same label (hence, sharing the same visual semantics) should lie closer to each other in the learned metric space more than those  images having different labels (see Fig. \ref{fig:triplet_loss}). 
In more detail,  a triplet ($\mathcal{T}$) of samples is sampled from  $\mathcal{G}$:   $\mathcal{T} = \{ (\mathbf{g}_{i}^{a}, \mathbf{g}_{i}^{p}, \mathbf{g}_{i}^{n})\}$, where $\mathbf{g}_{i}^{a}$ (called \textit{anchor}) is sampled randomly from  $\mathcal{G}$ together with its  associated class label $y_i$; $\mathbf{g}_{i}^{p}$ (called the  \textit{positive} sample) is drawn  among the samples having the same class labels ($y_i$) of the anchor; and $\mathbf{g}_{i}^{n}$ (called the \textit{negative} sample) is again chosen randomly from $\mathcal{G}$ such that it belongs to a class $y_j$, where $y_j \neq y_i$. For training the network with Stochastic Gradient Descent (SGD), a mini-batch of \textit{M}  triplets is sampled and the following {\em triplet loss} is minimized:
\vspace{-1mm}
\begin{equation}
	\label{eq.trip-loss}
	\begin{array}{cc}
		\mathcal{L}_{Metric} = &
		\displaystyle \sum_{i=1}^{M}max\Big(0, ||f(\mathbf{g}_{i}^{a}) - f(\mathbf{g}_{i}^{p})||_{2}^{2} - \\
		& ||f(\mathbf{g}_{i}^{a}) - f(\mathbf{g}_{i}^{n})||_{2}^{2} + \alpha \Big),
	\end{array}
\end{equation}
\noindent where $\alpha$ is the minimum {\em margin} threshold, which is forced between the positive and negative distances.

The network $f$ is composed of 3 fully-connected layers with  1024,  512 and  \textit{K} neurons each, being \textit{K} the number of desired hash bits. We use LeakyReLU in the two hidden layers to allow negative gradients to flow during the backward pass and a sigmoid activation in the final layer to restrict the output activations in [0, 1]. 

In order to push the final real activations towards the extremities of the sigmoid  range, and inspired by \cite{yang2018supervised}, we use a second loss  whose goal is to maximize the sum of the squared errors between the output-layer activations and the value 0.5: 
\vspace{-1mm}
\begin{equation}
\label{eq.push-loss}
\begin{array}{cc}
\mathcal{L}_{Push} = - \frac{1}{K} \displaystyle \sum_{i=1}^M || f(\mathbf{g}_i) - 0.5 \mathbf{1}||^2,
\end{array}
\end{equation}
\noindent where $\mathbf{1}$ is a vector of 1s having dimension \textit{K}.

Also inspired by \cite{yang2018supervised}, we adopt a third loss  which encourages each output neuron to fire with a 50\% chance. This means that the binary code representations of the images will have (on average) a balanced number of 0s and 1s, hence all the bits of the code are equally used. The balancing loss is:
\vspace{-1mm}
\begin{equation}
\label{eq.balan-loss}
\begin{array}{cc}
\mathcal{L}_{Balancing} =  \displaystyle \sum_{i=1}^M ( mean(f(\mathbf{g}_i)) - 0.5 )^2,
\end{array}
\end{equation}

\noindent where $mean(f(\mathbf{g}_i))$ is the mean of the output activations.

The final objective function is a weighted combination of the three losses:
\vspace{-1mm}
\begin{equation}
\label{eq.tot-loss}
\begin{array}{cc}
\mathcal{L} =  \mathcal{L}_{Metric} + \lambda_1 \mathcal{L}_{Push} + \lambda_2 \mathcal{L}_{Balancing}, 
\end{array}
\end{equation}

\noindent where $\lambda_1$ and $\lambda_2$ weight the relative loss importance.

When training of $f$ is done,  the final hashing function $h$ is obtained by quantizing the values in ${\rm I\!R}^K$. For a given archive image $x$, let   $\mathbf{g}$ be its intermediate feature obtained using Inception Net and $\mathbf{v}= f(\mathbf{g})$. The final binary code $\mathbf{b} = h(f(\mathbf{g}))$ is given by:
\vspace{-1mm}
\begin{align}
\label{thresholding}
\mathbf{b}_n &= (sign(\mathbf{v}_n - 0.5) + 1)/2, 1 \leq n \leq K
\end{align}

Finally, in order to retrieve an image $x_j$, which is semantically  similar to a query image $x_q$, we use the Hamming distance between $h(x_q)$ and  $h(x_j)$. 
Note that, with a slight abuse of notation, in the rest of the paper we write $h(x)$ to mean $h(f(\mathbf{g}))$, where $\mathbf{g}$ is the intermediate feature corresponding to $x$.
During retrieval, the Hamming distance is computed between the query $h(x_q)$ and each image in the archive and the obtained distances  are  sorted in ascending order of magnitude. The top-$k$ instances with the lowest distances are retrieved.

\section{Dataset Description And Design Of The Experiments}
In our experiments we  use two RS benchmarks. The first archive is the widely used UC Merced \cite{dai2018novel} (UCMD) archive that contains 2100 aerial images from 21 different land cover categories, where each category includes 100 images. The pixel size of each image in the archive is 256 $\times$ 256 and the spatial resolution is 30 cm. The second benchmark archive  is the Aerial Image Data Set \cite{xia2017aid} (AID) which is much larger with respect to the UCMD archive. The images are extracted from the Google Earth and and each image in the archive is a section of 600 $\times$ 600 pixels. AID contains 10,000 images from 30 different categories and the number of images in each category varies between between 220 to 420. The spatial resolution of the images  varies between 50 cm to 8 m.

\begin{table*}[t]
	\centering
	\caption{mAP@20 and average retrieval time in the UCMD archive.}
	\vspace{1mm}
	\begin{tabular}{|*{9}{c|}}
		\cline{4-9}
		\multicolumn{1}{c}{ } & \multicolumn{2}{c|}{ } & \multicolumn{6}{c|}{\# Hash Bits \textit{K}} \\
		\hline
		\multirow{2}{*}{Methods} & \multirow{2}{*}{mAP} & \multirow{2}{*}{Time (in ms)} & \multicolumn{2}{c|}{\textit{K}=16} & \multicolumn{2}{c|}{\textit{K}=24} & \multicolumn{2}{c|}{\textit{K}=32}\\
		\cline{4-9}
		& & & mAP & Time (in ms) & mAP & Time (in ms) & mAP & Time (in ms)\\
		\hline
		\multicolumn{1}{|c|}{\textbf{NN}-Inception feat. (Euclidean)} & 0.724 & 45.1 & - & - & - & - & - & -\\
		\hline
		\multicolumn{1}{|c|}{\textbf{NN}-Inception feat. (Hamming)} & 0.359 & 10.1 & - & - & - & - & - & -\\
		\hline
		\multicolumn{1}{|c|}{\textbf{KSLSH} \cite{demir2016hashing}} & - & - & 0.557 & 25.3 & 0.594 & 25.5 & 0.630 & 25.6\\
		\hline
		\multicolumn{1}{|c|}{Our \textbf{MiLaN}} & - & - & \textbf{0.875} & 25.3 & \textbf{0.890} & 25.5 & \textbf{0.904} & 25.6\\
		\hline
		\multicolumn{1}{|c|}{Our \textbf{MiLaN} (Eucledian)} & - & - & \textbf{0.903} & 35.3 & \textbf{0.894} & 35.8 & \textbf{0.916} & 36.0\\
		\hline
	\end{tabular}
	\label{tab:tab1}
\end{table*}

\begin{figure*}%
	\centering
	\subfloat[t-SNE scatter plot:  KSLSH \cite{demir2016hashing}\label{tsne_1}]{{\includegraphics[width=0.45\textwidth]{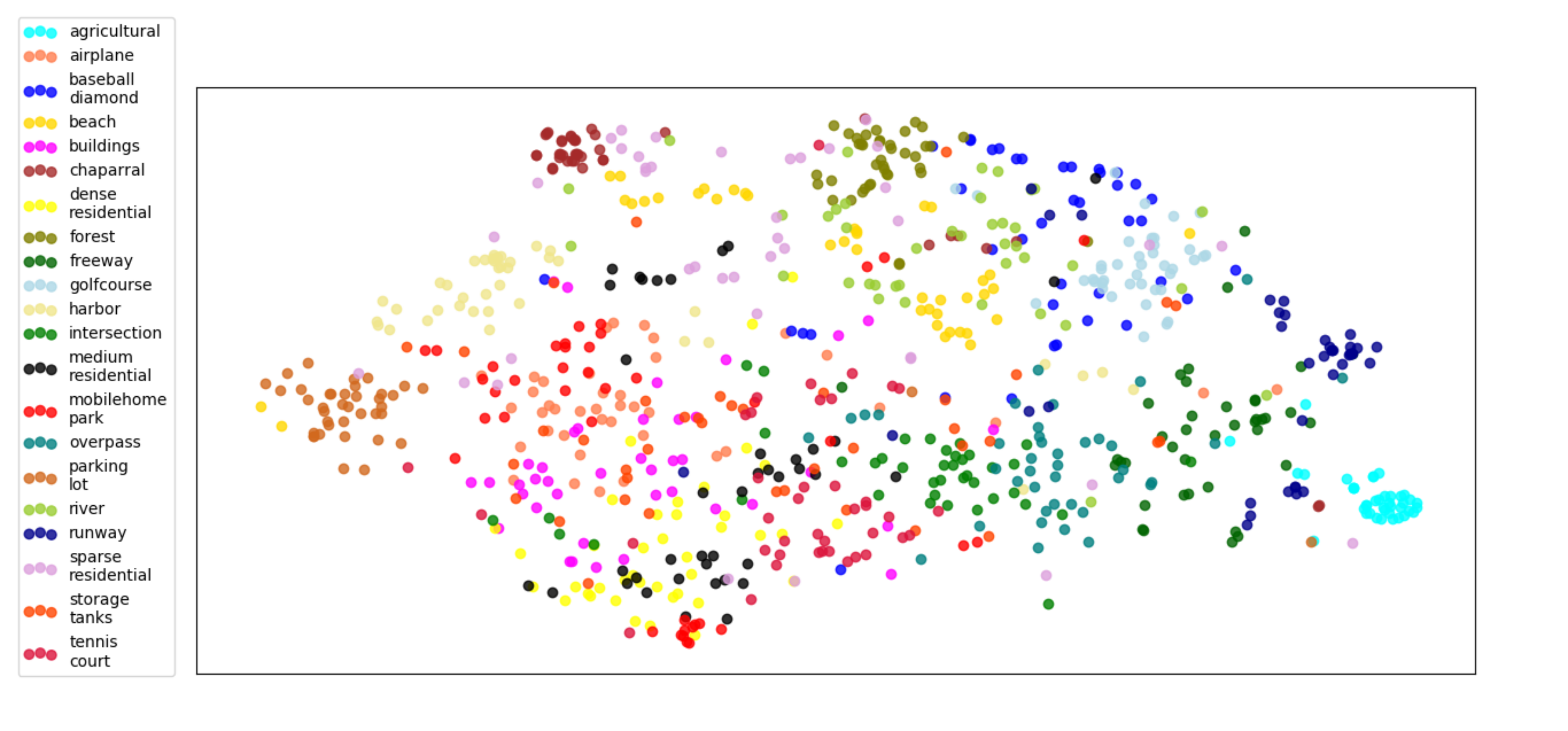} }}%
	\qquad
	\subfloat[t-SNE scatter plot:  MiLaN\label{tsne_2}]{{\includegraphics[width=0.45\textwidth]{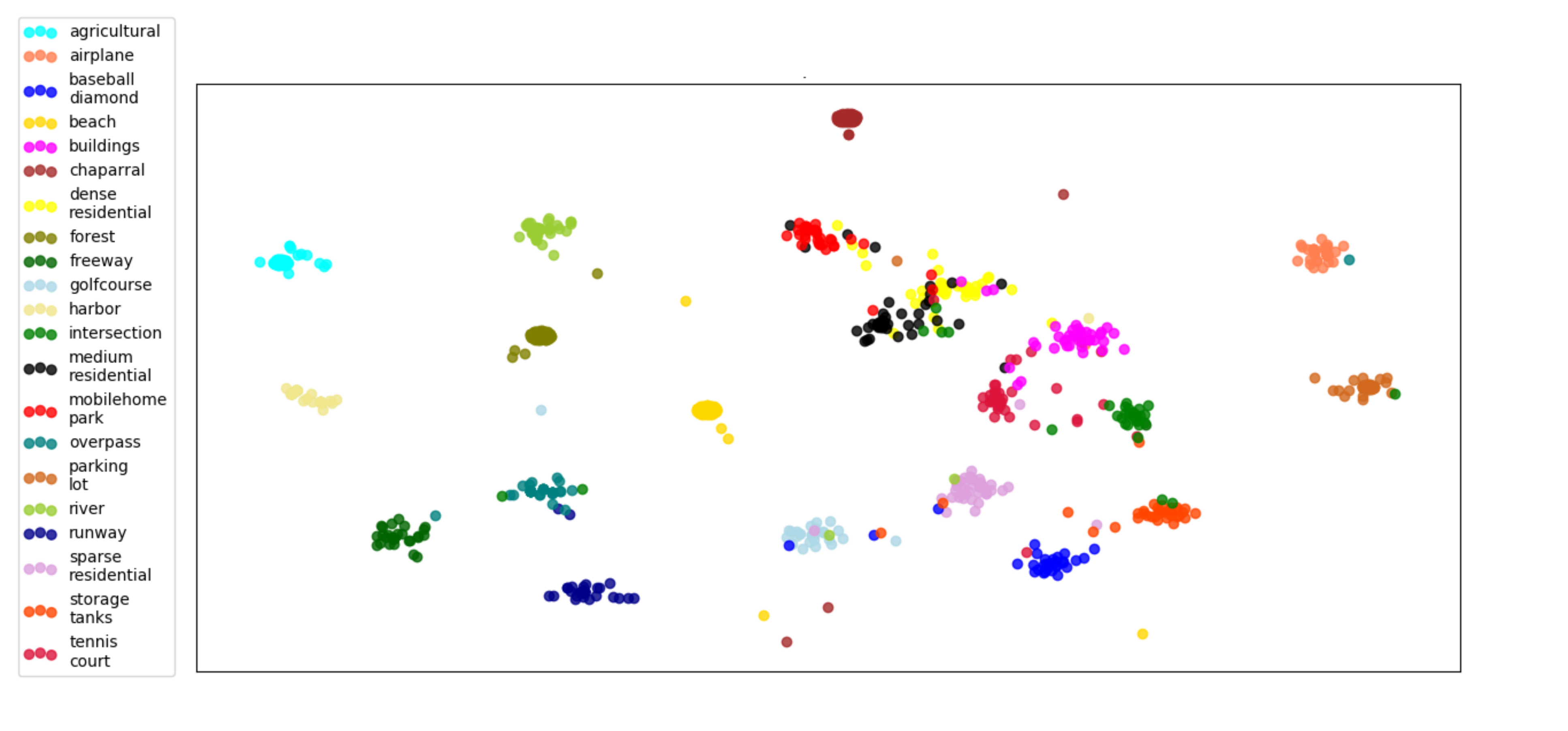} }}%
	\caption{t-SNE 2D scatter plots comparing the 2D-projection of the \textit{K}-dimensional binary hash codes of the test images in the UCMD archive. Each colour in the plot represents a land-cover category. This is best viewed in colour with maximum zoom.}%
	\label{fig:tsne}%
\end{figure*}

MiLaN\footnote{Our code is available at: \url{https://github.com/MLEnthusiast/MHCLN}} was trained using a mini-batch of triplets having a cardinality $M$=30. For a much larger archive we suggest to include only those triplets in the mini-batch, which contain semi-hard negatives (i.e., only those negative samples which violate the threshold margin $\alpha$) which allows for a faster convergence of the network. The value of the threshold margin $\alpha$ was set to $0.2$, which is determined by  cross-validation, using  a validation set composed of 20 images per class. The hyperparameters $\lambda_1$=0.001 and $\lambda_2$=1 have also been chosen using the same  validation set. We used the Adam Optimizer  with a small learning-rate $\eta$ = $10^{-4}$. The other two Adam hyper-parameters $\beta_1$ and $\beta_2$  were set to 0.5 and 0.9, respectively.

We compare MiLaN with the state-of-the-art supervised hashing methods 
KSLSH \cite{demir2016hashing} and DHNN \cite{li2018large}. However, due to the lack of a publicly available  code for DHNN, we can only report its results  for the UCMD archive.
A Gaussian RBF kernel is used for   KSLSH.
We also use, as baseline, a Nearest Neighbour (NN) search with our intermediate features. Specifically, 'NN-Inception feat. (Euclidean)' refers to a NN search using the Inception Net features and the Euclidean distance. The comparison with MiLaN shows the advantage of learning a metric space on top of these features. Moreover, 'NN-Inception feat. (Hamming)'
refers to a NN search {\em using binary hash codes} (of length $K = 2048$). These hash codes are computed using the same thresholding rule as in Eq. \ref{thresholding}.
The results of each method are provided in terms of: i) mean Average Precision (mAP); and ii) computational time. Specifically, mAP@\textit{k} is based on the top-\textit{k} retrieved images. We consider two different scenarios. In the first, no data augmentation is used and 60\% of all the images of each category is used for training and cross-validation, while the rest  for testing. In the second scenario, we use data augmentation obtained by means of basic geometric transformations.  

\vspace{-3mm}
\section{Experimental Results}

\subsection{Results: the UCMD Dataset}
\label{UCMDDatasetResults}
Table \ref{tab:tab1} shows that the  MiLaN mAP is significantly higher than 'NN-Inception feat. (Euclidean)' (for all the values of $K$), which shows the advantage of learning a metric space. Moreover, the retrieval time
is also drastically reduced by the fact that our network outputs binary hash codes and an efficient Hamming distance can be computed over the archive elements. On the other hand, 
'NN-Inception feat. (Hamming)' is faster (as expected) but the plain binarization of the CNN features leads to a {\em drastic} loss of information, which explains the results inferior to 'NN-Inception feat. (Euclidean)'. This shows the importance of {\em learning} how to compute effective binary codes, which in our case is the result of applying Eq.\ref{eq.push-loss} and Eq. \ref{eq.balan-loss}.
The last row of Table \ref{tab:tab1} reports the results obtained using the final MiLaN features {\em before} quantization (Eq. \ref{thresholding}). As expected, retrieval in an ${\rm I\!R}^K$ space is more effective than in $\{ 0,1 \}^K$. However, hash codes ('Our MiLaN') significantly improve  searching time, with a negligible accuracy drop. This shows that MiLaN is able to learn features which can be easily binarized while mostly preserving the metric space.

In order to show the contribution of the hash-code production losses, we report below the results obtained, respectively, with $K = 16, 24, 32$, when: (1) $\lambda_1 = \lambda_2 = 1$: 0.231, 0.524, 0.408 and  when: (2) $\lambda_1 = \lambda_2 = 0.001$: 0.823, 0.851, 0.907.

\begin{figure}[h!]
	\centering
	\includegraphics[width=0.9\columnwidth]{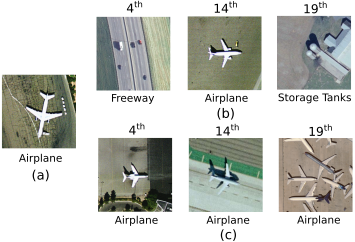}
	\caption{ (a) The query image. (b) Images retrieved by  KSLSH \cite{demir2016hashing}. (c) Images retrieved by our  MiLaN.}
	\label{fig:ret_imgs}
\end{figure}

 MiLaN, with \textit{K} = 24, also significantly outperforms (+29.6\% mAP)  KSLSH, with a  similar retrieval time. 
 Qualitative results are shown in Fig. \ref{fig:ret_imgs} where, using as query an image of an \enquote{airplane},  MiLaN retrieves semantically more similar images than  KSLSH.

We perform experiments on the UCMD archive also under the second scenario, where we consider data augmentation as suggested by \cite{li2018large}. Accordingly, we replicate the experimental set-up of  DHNN \cite{li2018large} by augmenting the original data set with images rotated by 90$^{\circ}$, 180$^{\circ}$ and 270$^{\circ}$, yielding 8400 images. Then, out of the 8400 images, 7400 are randomly chosen for training the network and remaining 1000 images are used for evaluation. It is worth noting that the 7400 training images are also used as a search set as suggested by \cite{li2018large}. Table \ref{tab:tab3} compares  DHNN \cite{li2018large} with our MiLaN using mAP. These results show that  MiLaN requires a smaller \textit{K} to reach a higher mAP. For instance, when \textit{K} = 32, the mAP for  MiLaN is 3.8\% higher than that of  DHNN. Moreover, the mAP of MiLaN with \textit{K} = 32 is again 0.6\% higher than the mAP obtained by  DHNN with twice the number of hash bits, i.e., \textit{K} = 64. We highlight that the use of such an experimental set-up, as proposed by \cite{li2018large}, may lead to the identical but rotated variants of a same image being present in both the training and the test sets and thus leading to over-saturated results. In fact, the network can memorize the test samples (rotated variants) which are present in the training set. However, we adopt this training and evaluation protocol in order to fairly compare with \cite{li2018large}.

\begin{table}[h]
	\centering
	\def\arraystretch{1.2}
	\setlength{\tabcolsep}{12pt}
	\caption{mAP with data augmentation for the UCMD archive.}
	\begin{tabular}{|*{4}{c|}}
		\hline
		& DPSH~\cite{li2015feature} & DHNN~\cite{li2018large} & Our \textbf{MiLaN} \\
		\hline
		\textit{K} = 32 & 0.748 & 0.939 & \textbf{0.977} \\
		\hline
		\textit{K} = 64 & 0.817 & 0.971 & \textbf{0.991} \\
		\hline
	\end{tabular}
	\label{tab:tab3}
\end{table}

\begin{table*}[h]
	\centering
	\caption{mAP@20 and average retrieval time  for the AID archive.}
	\vspace{1mm}
	\begin{tabular}{|*{9}{c|}}
		\cline{4-9}
		\multicolumn{1}{c}{ } & \multicolumn{2}{c|}{ } & \multicolumn{6}{c|}{\# Hash Bits \textit{K}} \\
		\hline
		\multirow{2}{*}{Methods} & \multirow{2}{*}{mAP} & \multirow{2}{*}{Time (in ms)} & \multicolumn{2}{c|}{\textit{K}=16} & \multicolumn{2}{c|}{\textit{K}=24} & \multicolumn{2}{c|}{\textit{K}=32}\\
		\cline{4-9}
		& & & mAP & Time (in ms) & mAP & Time (in ms) & mAP & Time (in ms)\\
		\hline
		\multicolumn{1}{|c|}{\textbf{NN}-Inception feat. (Euclidean)} & 0.719 & 145.1 & - & - & - & - & - & -\\
		\hline
		\multicolumn{1}{|c|}{\textbf{NN}-Inception feat. (Hamming)} & 0.402 & 60.2 & - & - & - & - & - & -\\
		\hline
		\multicolumn{1}{|c|}{\textbf{KSLSH} \cite{demir2016hashing}} & - & - & 0.426 & 115.3 & 0.467 & 116.1 & 0.495 & 117.5\\
		\hline
		\multicolumn{1}{|c|}{Our \textbf{MiLaN}} & - & - & \textbf{0.876} & 117.5 & \textbf{0.891} & 116 & \textbf{0.926} & 114.5\\
		\hline
	\end{tabular}
	\label{tab:tab2}
\end{table*}

\begin{figure*}[!ht]
	\subfloat[\label{abl_1}]{%
		\includegraphics[width=0.25\textwidth]{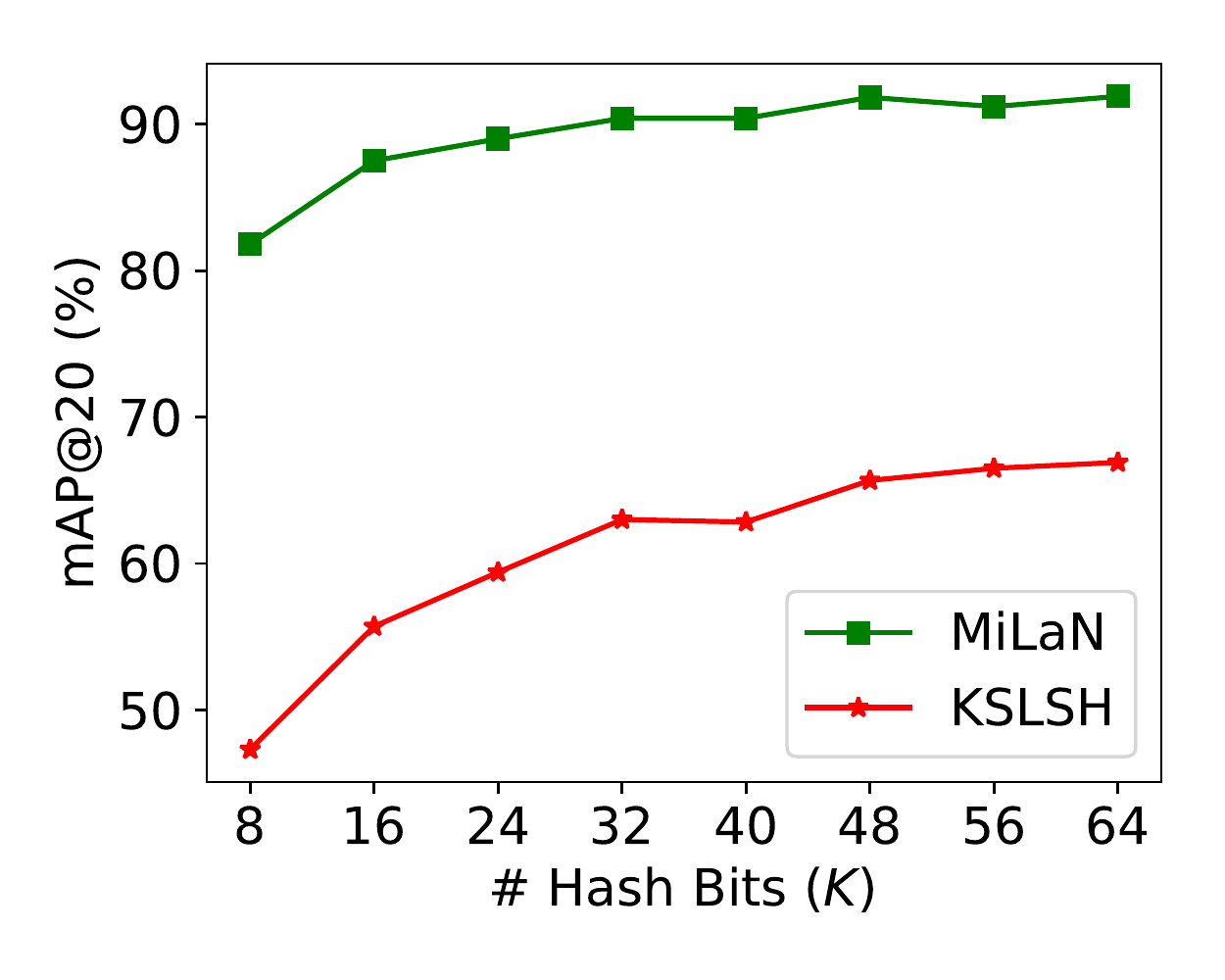}
	}
	\subfloat[\label{abl_2}]{%
		\includegraphics[width=0.25\textwidth]{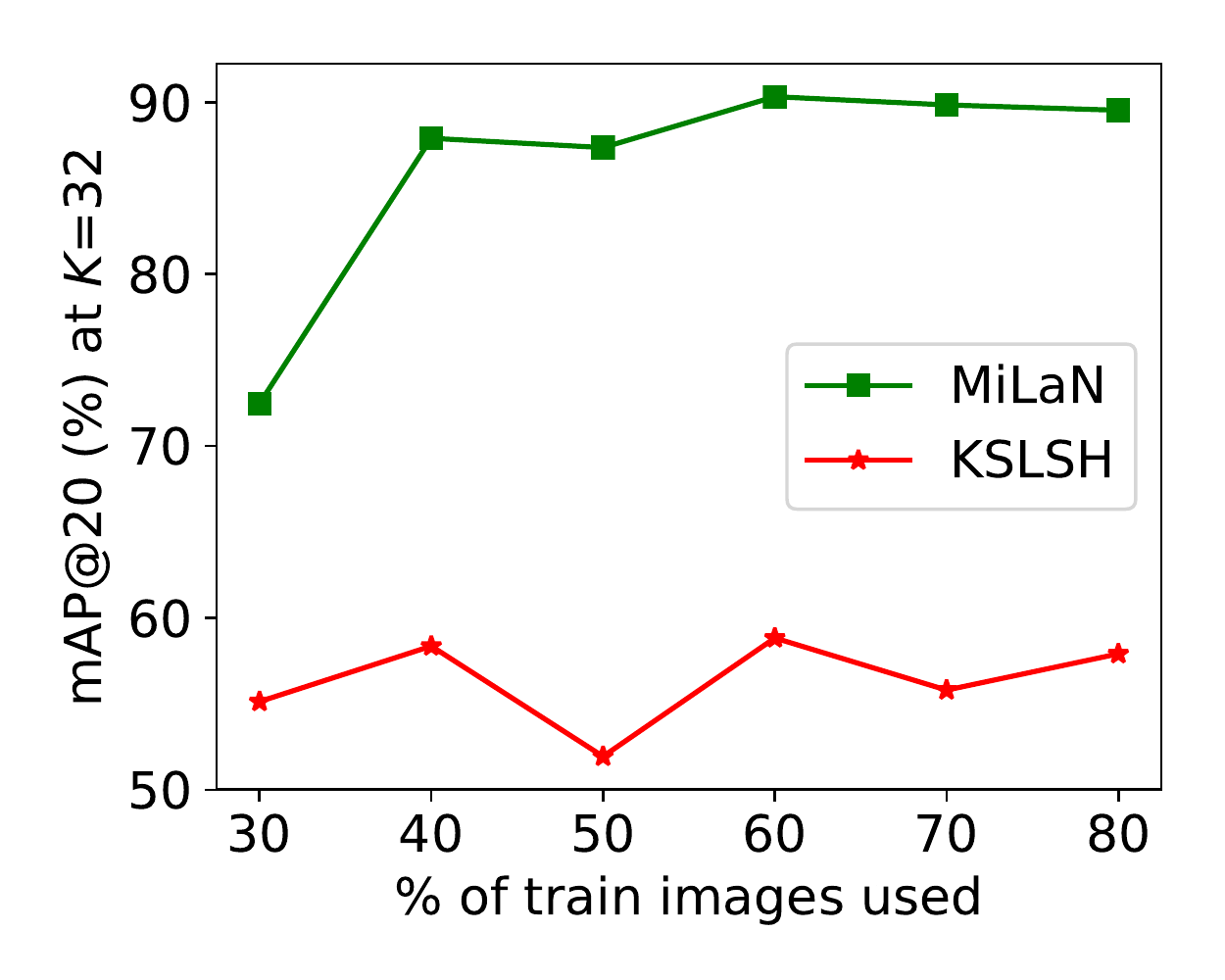}
	}
	\subfloat[\label{abl_3}]{%
		\includegraphics[width=0.25\textwidth]{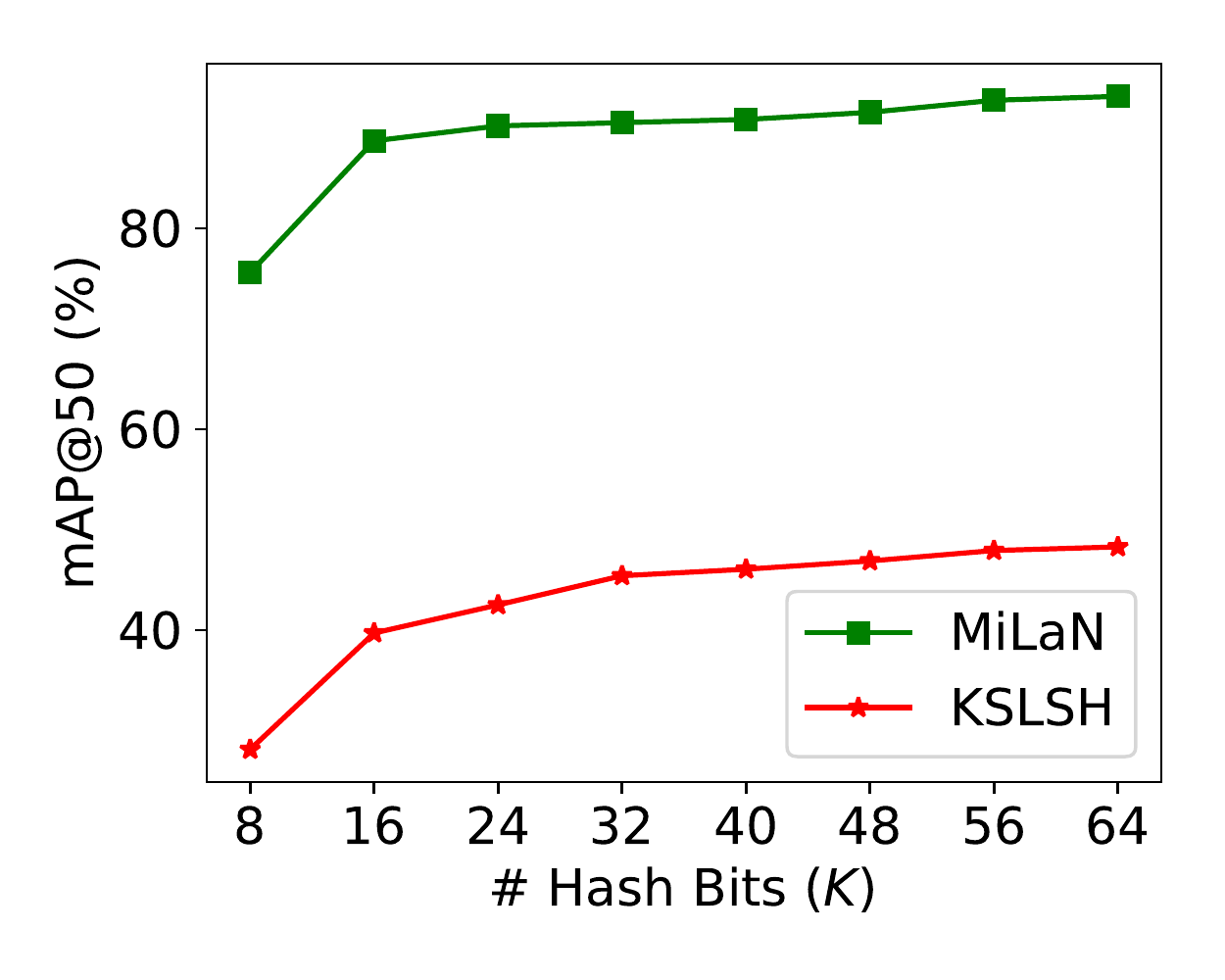}
	}
	\subfloat[\label{abl_4}]{%
		\includegraphics[width=0.25\textwidth]{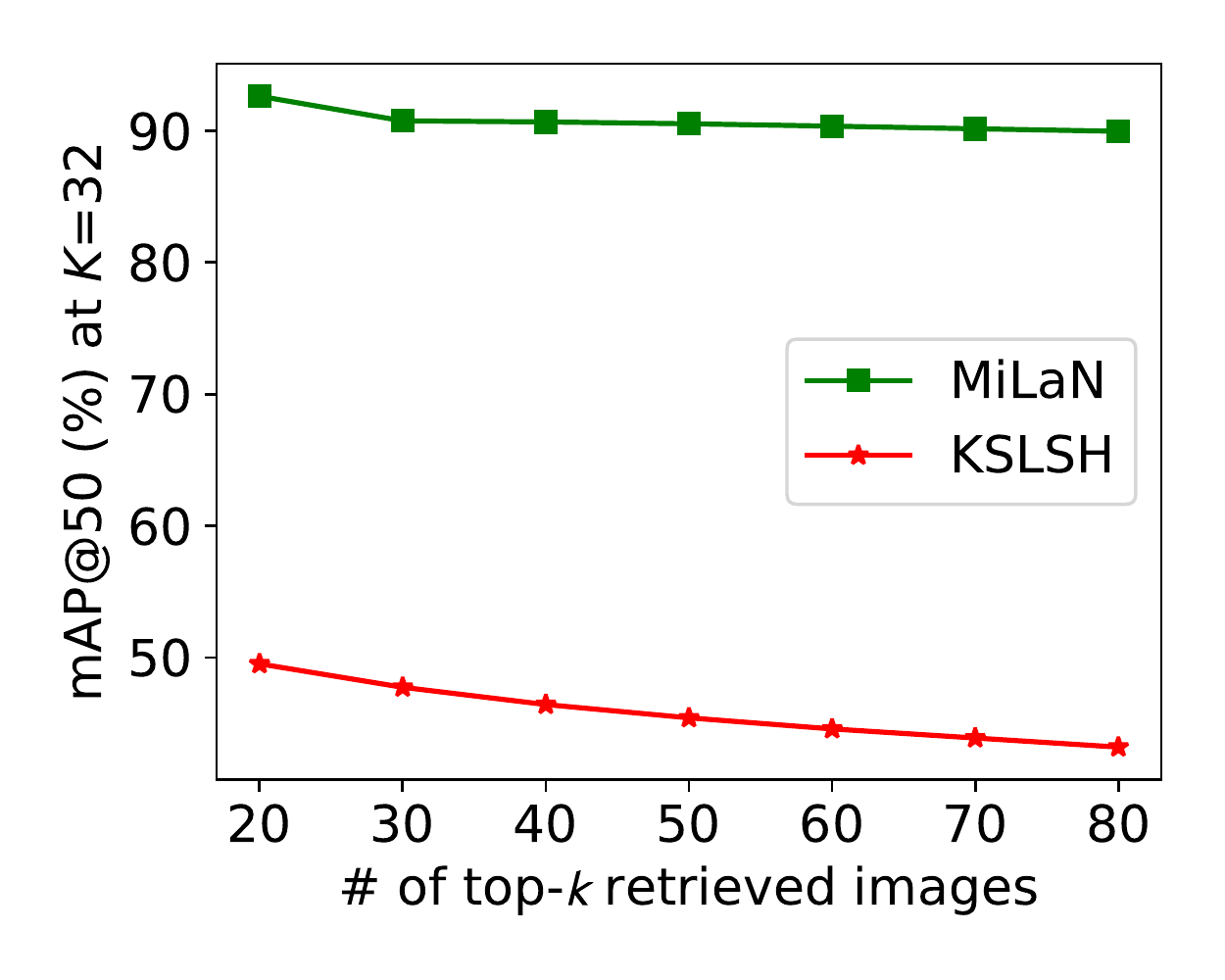}
	}
	\caption{Comparative analysis of  KSLSH and  MiLaN. (a) and (b): mAP@20 curves of  KSLSH and  MiLAN on  UCMD for (a) different number of hash bits; and (b) different train:test ratio at $K=32$. (c) and (d): mAP@50 curves of  KSLSH and  MiLAN on  AID for (c) different number of hash bits; and (d) different number of top-$k$ retrieved images at $K=32$.}
	\label{fig:abl}
\end{figure*}

To visualize the \textit{K}-dimensional binary codes corresponding to the test images, we perform a t-distributed stochastic neighbour embedding (t-SNE) which projects these representations in a 2D space. Fig. \ref{tsne_1} shows the results in case of KSLSH. Note that, although  the samples of the \enquote{agricultural} category stand out clearly from the rest of the samples belonging to the other categories, the remainder of the samples are mostly cluttered in one big heap having minimal inter-class separation. This shows that  KSLSH is not discriminative enough. In contrast, Fig. \ref{tsne_2} shows that  MiLaN leads to samples exhibiting much more compact clusters. For instance, the categories  \enquote{freeway} and \enquote{overpass}, which exhibit similar visual patterns, lie close to each other and yet far enough to have their own well-separated clusters.

We  also analyse the impact of the length of the hash codes on the retrieval performance.  Fig. \ref{abl_1} shows that,  both for KSLSH and for MiLaN, the mAP@20  increases with an increase in the number of the hash bits. However, the average precision curve for MiLaN with respect  to all the \textit{K} values, stays considerably above the KSLSH results. This indicates that the proposed network efficiently maps semantic information into its discriminative hash codes with varying lengths, significantly outperforming  the corresponding KSLSH results. Finally, we train MiLaN with different train:test splits and report  mAP@20 at \textit{K}=32 in Fig. \ref{abl_2}. For instance, Fig. \ref{abl_2} shows that the mAP obtained by  MiLAN using only 30\% of the training images is much higher than the mAP obtained by  KSLSH with 80\% of the total  training images.

\subsection{Results: the AID Dataset}
In the experiments  on the AID archive, we split the images of each category with a 60:40 train:test split-ratio. The results of this archive demonstrate similar trends as those in  UCMD. In summary,  MiLaN achieves a higher mAP@20 with respect to all the other methods (Table \ref{tab:tab2}). For instance, with \textit{K} = 16, the mAP@20 of MiLaN is 45\% higher than that of  KSLSH. A similar trend is kept between these two methods for higher values of \textit{K}. 
Moreover,  the AID dataset results confirm all the observations pointed out in Sec. \ref{UCMDDatasetResults} concerning the comparison among MiLaN,'NN-Inception feat. (Euclidean)' and 'NN-Inception feat. (Hamming)', again confirming the advantage of simultaneously learning a metric space and the hash codes on top of the adopted CNN features.

Finally, we report an ablation study conducted on the AID archive. Fig. \ref{abl_3} shows that, with increasing values of $K$, the map@50 values of MiLaN increases monotonically and outperforms  KSLSH for all values of $K$. Similarly, in Fig. \ref{abl_4}, we observe that, when the number of the top-$k$ retrieved images increases, the map@50 value  slightly decreases for MiLaN but quite significantly for the KSLSH.

\section{Conclusion}
In  this  letter  we  have  introduced  a  metric-learning  based deep  hashing  method  for  fast  and  accurate  RS CBIR.  The proposed approach is based on  an intermediate representation  of the RS images obtained using an external, not-retrained CNN (in our case we used Inception Net, pre-trained on ImageNet, but other networks can be used as well). This representation is important to avoid overfitting risks in small-size labeled RS datasets. Using this representation as input instead of raw-pixel data, we train our hashing network using three different losses: a triplet loss which is in charge of  learning a  metric space and a balancing and a representation loss which  effectively  learn to produce binary  hash  codes. 
In this way we jointly solve two problems: we avoid overfitting risks using the intermediate features and we learn how to produce similarity-based hash codes. Our empirical analysis shows the advantage of this combined strategy, which is more accurate and  time-efficient than  state-of-the-art methods. 


\section*{Acknowledgments}
This work was supported by the European Research Council (ERC) through the ERC-2017-STG BigEarth Project under Grant 759764.

\vspace{-2mm}


%


\ifCLASSOPTIONcaptionsoff
  \newpage
\fi



%

\bibliographystyle{IEEEtran}
\bibliography{references}

%





\end{document}